\documentstyle[IPMU96]{article}

\title{
On Marginally Correct Approximations 
of Dempster-Shafer
 Belief Functions from Data
}

\newcommand{\SQL}[1]{
\begin{description}
\item{}{\it #1}
\end{description}
}
\date{}
\author{ {\bf Mieczys{\l}aw A. K{\l}opotek} and {\bf S{\l}awomir T. Wierzcho\'{n}} \\  
Institute of Computer Science\\ 
Polish Academy of Sciences\\
01-237 Warsaw, ul. Ordona 21, Poland\\
E-mail: klopotek{@}ipipan.waw.pl,  stw{@}ipipan.waw.pl
} 
\begin{document}
\maketitle
\begin{abstract}
Mathematical Theory of Evidence (MTE)
is blamed to leave frequencies outside
its framework. 
%
In this paper we consider this problem from the point of view 
of conditioning in the MTE. We describe the class of belief functions 
for which 
 marginal consistency with observed frequencies may be achieved and 
 conditional belief functions are proper belief functions,
and deal with implications for 
approximation of general 
belief
functions by this class 
 and
for inference models in MTE. 

\end{abstract} 
\section{INTRODUCTION}
The Dempster-Shafer Theory  or the Mathematical Theory of Evidence (MTE) 
\cite{Shafer:76,Dempster:67} 
 is intended to be a generalization of bayesian
theory of subjective probability \cite{Shafer:90ijar}. 
This theory offers  capability of
representing ignorance in a simple and direct way, compatibility with the
classical probability theory, compatibility with boolean logic and feasible
computational complexity  \cite{Ruspini:92ijar}. 
MTE may be applied for (1) representation of incomplete knowledge, (2) belief
updating, (3) and for combination of evidence  \cite{Provan:92}.
MTE covers the statistics of random sets and may be applied for representation
of incomplete statistical knowledge. Random set statistics is quite popular in
analysis of opinion polls whenever partial indecisiveness of respondents is
allowed \cite{Dubois:92}. 
%
Practical applications of MTE include: integration of knowledge from
heterogeneous sources for object identification  \cite{deKorvin:93}, 
technical diagnosis under unreliable measuring devices  \cite{Durham:92}, 
medical applications: \cite{Gordon:90,Zarley:88b}.


In spite of indicated merits, MTE experienced sharp criticism from many sides.
The basic line of criticism is connected with the relationship between the
belief function (the basic concept of MTE) and frequencies
\cite{Wasserman:92ijar,Halpern:92}. 
 A number of attempts to interpret belief functions in terms of probabilities
have failed so far to produce a fully compatible interpretation with MTE - see
e.g. \cite{Kyburg:87,Halpern:92,Fagin:91} etc. Shafer
\cite{Shafer:90ijar} and Smets \cite{Smets:92}, in defense of MTE, dismissed
every attempt to interpret MTE frequentistically. Shafer stressed that
even modern (that meant bayesian) statistics is not frequentistic at all
(bayesian theory assigns subjective probabilities), hence frequencies be no
matter at all.\\
Wasserman \cite{Wasserman:92ijar} strongly opposed claims of Shafer
\cite{Shafer:90ijar} about frequencies and bayesian theory. Wasserman pointed
out that the major success story of bayesian theory is  the
exchangeability theory of  de Finetti, which  treats frequency based
probabilities as a special case of bayesian belief. Hence frequencies, as
Wasserman claims, are inside the bayesian theory, but outside the
Mathematical Theory of Evidence.

In this paper we consider this problem from the point of view 
of apriorical and aposteriorical conditioning in the MTE. We describe the
class of belief functions for which: 
 marginal consistency with observed frequencies may be achieved, 
apriorical conditional belief functions are proper belief functions

 {
We will assume that the Reader is familiar with basic concepts of MTE like:
belief function (Bel), basic probability assignment function (bpa, or m),
commonality function(Q), marginalization (projection) onto a subset s of the
set of all variables ($Bel ^{\downarrow s}$), vacuous extension onto the set
of variables ($Bel ^{\uparrow s}$), focal points (sets for which bpa is
non-zero). These terms are explained
in many standard papers and books on MTE \cite{Shafer:76,Shenoy:90}. \\
}

\section{CASE-BASED UNDERSTANDING OF BELIEF FUNCTIONS}

In an attempt to overcome the reason for those numerous failures to
interpret consistently
Dempster's rule of combination, 
a new frequency interpretation of MTE has been proposed in
\cite{Klopotek:94e}. It has been demonstrated there that, in general, 
Dempster's rule has a destructive impact on the data so that 
whatever we expect
calculation of conditional probabilities (given an event) differs from
whatever we obtain under conditioning (given an event) in MTE (see the
definition of aposteriorical conditioning of Shafer in \cite{Shafer:90}.

Let us have a look at SQL construct to calculate joint probability
distribution from a database in two variables P(X,Y):
%
\SQL{
create view Total (Counted)  as select count(*) from Cases;
}
\SQL{
create view Probability (X,Y,Prob) as 
select X,Y,\\
count(*)/Counted from Cases,Total group by X,Y;
}

Let us look at calculation of conditional probability $P(X,Y|X \in A_X)$. 
We proceed as follows: we select first the proper subset of cases from the
database and proceed to calculate the unconditional probability
for the selected cases.
%
\SQL{
create view SelectedCases (X,Y)
as select X,Y
 from Cases  where
$X \in AX$;
}
\SQL{
create view TotalCond (Counted) as select count(*) from SelectedCases;
}
\SQL{
create view CondProbability (X,Y,CondProb) as 
select Y,X,count(*)/Counted \\
from SelectedCases,TotalCond  where $X \in AX$
group by X, Y;
}

If we calculate a conditional probability       
$P(X,Y|X \in A_X,X \in B_X)$ on a series of conditions $(X \in A_X,X \in B_X)$
we can proceed first by selecting cases for  the first, then for the second
condition etc., finding the intersection, and then calculating the
probabilities (id - the identifier).
\SQL{
create view SelectedCasesA (id,X,Y)
as select id,X,Y
 from Cases  where
$X \in AX$;
}
\SQL{
create view SelectedCasesB (id,X,Y)
as select id,X,Y
 from Cases  where
$X \in BX$;
}
\SQL{
create view SelectedCases (id,X,Y)
as select id,X,Y 
 from SelectedCasesA \\
 intersection 
   select id,X,Y 
 from SelectedCasesB ;
}
\SQL{
create view TotalCond (Counted) as select count(*) from SelectedCases;
}
\SQL{
create view CondProbability (X,Y,CondProb) as 
select Y,X,count(*)/Counted \\
from SelectedCases,TotalCond  where $X \in AX$
group by X, Y; 
}

Let us look at calculation of basic probability assignment (bpa)
m(X,Y) 
 from data
 :
\SQL{
create view Total (Counted) as select count(*) from Cases;
}
\SQL{
create view bpa (X,Y,m) as 
select X,Y,\\
count(*)/TCounted from Cases,Total group by X,Y;
}

Let us look at calculation of aposterioric
 conditional bpa $m(X,Y||X \in A_X)$
We have to proceed as follows: we select first the proper subset of cases
from the database and
MODIFY the values for the variable X. Then we 
 proceed to calculate the unconditional bpa        
for the selected and updated cases.
%
%
%
\SQL{
create view UpdatedCases (X,Y) as 
select $X \cap AX$,Y from Cases  where $X \cap AX\ne
\emptyset$;
}
\SQL{
create view TotalCond (Counted) as select count(*) from UpdatedCases;
}
\SQL{
create view Cond\_bpa (X,Y,Condbpa) as 
select X,Y,count(*)/Counted \\
from UpdatedCases,TotalCond  ;
}

If we calculate a conditional bpa 
$m(||X \in A_X,X \in B_X)$ on a series of conditions $(X \in A_X,X \in
B_X)$
we CANNOT proceed  by selecting cases for  the first, then for the second
condition etc., finding the intersection, and then calculating the
bpa, because this would yield wrong results due to side effects stemming from 
 case modifications.

 This has a serious
impact if we try to factorize a belief function into simpler components, e.g.
for purposes of propagation of uncertainty (methods of propagation of
uncertainty are presented e.g. in \cite{Cano:93,Shenoy:90}.
 It turns out
that:
\begin{itemize} 
\item It is, in general, impossible to factorize a joint belief distribution
into components being conditional belief functions
\footnote{Notice that in the domain of probability distributions, EVERY 
probability distribution may be represented by a composition of conditional
probability distributions as so-called bayesian network}
 (see eg. \cite{Cano:93} for
a discussion why two different notions of conditioning are needed for MTE:
the posteriori-conditionals as conditionals in the sense of Shafer, and
a priori-conditionals invented by Cano et al.,)
\item A priori-conditional belief functions as proposed by Cano et al. 
in general do not exists (see a discussion on
non-existence of a-priori conditionals in MTE presented by \cite{Shenoy:94})
\footnote{In probability distributions, a-priori and a-posteriori
conditionals
from the point of view of frequencies
 coincide, and they always exist:
Let us assume P(X,Y) is a frequency based (case-based) probability
distribution of X,Y. A posteriori conditional distribution of variable X 
given variable Y=y
 is interpreted as 
uncertainty distribution  of variable X $P'(X|Y=y)$ if we restrict ourselves
only to cases for which Y=y. Apriori conditional distribution of X given Y is
a
function $P(X|Y)$ in X,Y defined as $P(X|Y)=P(X\cap Y)/P(Y)$ for all values of
y for which $P(Y=y)>0$. Obviously: $P'(X=x|Y=y)=P(X=x|Y=y)$ if $P(Y=y)>0$.
} 
\item  What is more, it is often impossible to factorize a belief function
$Bel$
in variables $p,q,r$ into two factors one $Bel_1$ in variables  $p,r$ and the
second $Bel_2$ in variables $q,r$ even if in conditional distribution given
any value of the variable $r$, variables $p,q$ are independent (see
\cite{Studeny:94}, that is when for
every subset $r_i$ of the domain of the variable r the following holds: %
 $$Bel(||r_i)^{\downarrow p,q}=  
 Bel(||r_i)^{\downarrow p} \oplus 
 Bel(||r_i)^{\downarrow q}$$
 \end{itemize}

Therefore in papers \cite{Klopotek:93f,
Klopotek:94b,
Klopotek:94e} we have presented another approach to
factorization of belief functions in terms of anticonditional belief
functions. It turns out, however, that   anticonditional belief
functions are in general not belief functions but only pseudo-belief functions
(that is ones with non-negative commonality functions). Thus,         
anticonditional belief
functions have no direct counterparts in the physical world, as the basic
probability assignment may take negative values.
 
One can be tempted to suggest, that one shall  then  resign  from 
modeling the
joint belief distribution and instead try to find a marginally consistent 
decomposition of the joint belief distribution.  But:

\begin{itemize}
\item What is the class of Dempster-Shafer (DS) belief functions for which
apriori-condi\-tion\-al  
belief functions exist ?
\item How can general belief functions and uncertainty propagation for them
be approximated by this class of belief functions   and uncertainty
propagation for them ?
\item How can the belief functions and the reasoning with them be related to
frequencies (cases)?
\end{itemize}

\section{CORRECT APPROXIMATION OF CONDITIONALS}

Shafer \cite{Shafer:90} suggested that a belief function may emerge if
we observe a variable $X$ with domain of values $\Xi_X$ indirectly (via a
mapping f)
by observing actually another variable Y with a domain $\Xi_Y$ such that the
mapping $f:\Xi_Y\rightarrow\Xi_X$ is not a function. Hence, in some cases, if
we observe a single value $\xi_Y$ of the variable Y, we can tell at most that
the value of the variable X belongs to a non-empty set of elements
 \{$\xi_{X_1},
\xi_{X_2},\dots,\xi_{X_n}$\}. So a probability distribution P(Y) in $Y$
translates
into a basic belief assignment function, from which one calculates easily
belief distribution $Bel
^X$ in X. In this way we could get a case-based distribution in variable  
X.
\begin{center}
\begin{tabular}{|c|c|r|r|r|}
\hline
\multicolumn{1}{|p{1cm}|}{Variable Y}& 
\multicolumn{1}{p{2cm}|}{Corresponding set values of X (A)} 
&
\multicolumn{1}{p{0.8cm}|}{fre\-quen\-cy}
& \small$ m
^X(A)$&\small$Bel ^X(A)$\\
\hline 
$y_1$      & \{$x_1$\}      &     10   &   .10       &  .10 \\
$y_2$      & \{$x_1,x_2$\}  &     20   &   .20       &  .30 \\
$y_3$      & \{$x_2,x_3$\}  &     30   &   .30       &  .70 \\
$y_4$      & \{$x_3$\}      &     40   &   .40       &  .40 \\
--         & \{$x_1,x_2,x_3$\}  & --   &   .0        & 1.0 \\
\hline 
\end{tabular}
\end{center}
This could be extended simply to multivariate
belief distributions. However, we encounter one interpretational problem:\\
Let us consider the following frequency table in variables X and Z:
\begin{center}
\begin{tabular}{|l|l|r|}
\hline
X & Z & frequencies\\
\hline
\{$x_1,x_2$\} & \{$z_1,z_2\}$ & 20\\
\hline
\end{tabular}
\end{center}
What shall be the focal points of the multivariate belief function $Bel ^{X,
Z}$ in two variables X,Z~? Marginal consistency will be achieved
either 
 if we
assume $$m_1 ^{X,
Z}(\{(x_1,z_1), (x_1,z_2), (x_2,z_1), (x_2,z_2)\})=1$$ and if 
 we
take   $$m_2 ^{X,
Z}(\{(x_1,z_1), (x_2,z_2)\})=1$$ and if 
 we
suppose  $$m_3 ^{X,
Z}(\{ (x_1,z_2), (x_2,z_1)\})=1$$ and for many other
belief functions. But which one is the best~?
Let us try to calculate the conditional belief function $Bel_i(||X=x_1)
^{\downarrow Z}$. (From Shafer's formula:  $Bel_i(||X=x_1)
^{\downarrow Z} = (Bel_i \oplus Bel_{X=x_i})^{\downarrow Z} $ with
$Bel_{X=x_i}$ being a belief function with the only focal point 
$m_{X=x_i}(\{(x_1,z_1), (x_1,z_2)\})=1$).  We  get  three  totally 
different
results - three belief functions with differing focal points:
$m_1(||X=x_1)
^{\downarrow Z}(\{z_1,z_2\})=1 $, 
$m_2(||X=x_1)
^{\downarrow Z}(\{z_1\})=1 $, 
$m_3(||X=x_1)
^{\downarrow Z}(\{z_2\})=1 $. 

It is next to impossible to decide which of these
conditionals is the correct one from the point of view of the observed data.
So  which  belief  function  shall  be  treated  as   the    most 
representative
for the data ?
 We suggest here the first one ($Bel_1$) for the following reasons:
\begin{enumerate}
\item 
 if we observe $X,Z$ separately, we have no reason to assume that $x_1$ must
co-occur with $z_1$ but never with $z_2$ etc - we assume we have no more
information than actually visible from the data,
\item
 the methods of uncertainty propagation suggested both by Cano et al.
\cite{Cano:93} and Shenoy \& Shafer \cite{Shenoy:90} implicitly assume that
the joint belief in values of observed variables $X_1,X_2,...X_n$ is the
composition of the values of individual variables:
$$Bel_{observ}=Bel_{X_1=A_1} \oplus Bel_{X_2=A_2} \oplus \dots \oplus
Bel_{X_n=A_n}$$
with $A_i\subseteq \Xi_{X_i}$ being a subset of the domain of the ${i}^{th}$
variable.  Violation of this assumption would invalidate the respective method
of uncertainty propagation.
\end{enumerate}

Let us consider now the following data in three (logical) variables X,Y,Z,
giving the belief function $Bel_{and}$. 
\begin{center}
\begin{tabular}{|c|c|c|r|r|}
\hline
X= 
&  Y= 
& Z=
&
\multicolumn{1}{p{0.8cm}|}{fre\-quen\-cy}& 
\multicolumn{1}{p{0.8cm}|}{$m_{and}$ $(A_X\times A_Y\times A_Z)$}\\
\hline 
$\{t_X   \}$    & $\{t_Y   \}   $ & $ \{t_Z   \}  $ &   10       &  .10 \\
$\{t_X   \}$    & $\{f_Y   \}   $ & $ \{f_Z   \}  $ &   10       &  .10 \\
$\{f_X   \}$    & $\{t_Y   \}   $ & $ \{f_Z   \}  $ &   10       &  .10 \\
$\{f_X   \}$    & $\{f_Y   \}   $ & $ \{f_Z   \}  $ &   10       &  .10 \\
$\{t_X   \}$    & $\{t_Y,f_Y \} $ & $ \{t_Z,f_Z \}$ &   10       &  .10 \\
$\{f_X   \}$    & $\{t_Y,f_Y \} $ & $ \{f_Z   \}  $ &   10       &  .10 \\
$\{t_X,f_X \}$  & $\{t_Y   \}   $ & $ \{t_Z,f_Z \}$ &   10       &  .10 \\
$\{t_X,f_X \}$  & $\{f_Y   \}   $ & $ \{f_Z   \}  $ &   10       &  .10 \\
$\{t_X,f_X \}$  & $\{t_Y,f_Y \} $ & $ \{t_Z,f_Z \}$ &   20       &  .20 \\
\hline 
\end{tabular}\\
\end{center}
Let us consider calculation of an apriori conditional belief function in the
sense of Cano et al. \cite{Cano:93} such that it would imply the value of
Z given X,Y. A look at the data would suggest that variables X,Y and Z are
connected by the logical equation: $X \land Y = Z$ so that one might suggest 
a belief function with focal point $$m_\&( \{
(t_X,t_Y,t_Z),
(t_X,f_Y,f_Z),$$ $$
(f_X,t_Y,f_Z),
(f_X,f_Y,f_Z)
\})=1$$ is the apriori conditional connecting X,Y and Z. But this is a wrong
conclusion, because:
$$Bel_{and} \neq Bel_{and}^{\downarrow \{X,Y\}} \oplus Bel_\&$$
What is more, the Cano et al a priori conditional of Z given X,Y does not
exist at all !!! Therefore, decomposition of a joint belief distribution into
conditionals cannot in general be achieved. 
However, we must acknowledge that marginal consistency is actually achieved,
that is :
$$(Bel_{and})^{\downarrow X} 
 =  (Bel_{and}^{\downarrow \{X,Y\}} \oplus Bel_\&)^{\downarrow X} $$
$$(Bel_{and})^{\downarrow Y} 
 =  (Bel_{and}^{\downarrow \{X,Y\}} \oplus Bel_\&)^{\downarrow Y} $$
$$(Bel_{and})^{\downarrow Z} 
 =  (Bel_{and}^{\downarrow \{X,Y\}} \oplus Bel_\&)^{\downarrow Z} $$

Can we always construct a marginally consistent apriori-conditional~? We shall
assume that we have found a marginally consistent apriori conditional $Bel
^{|p}$ of Belief function Bel given set of variables p iff $Bel
^{|p}$ is a belief function and  for every variable X:
$$(Bel)^{\downarrow X}= (Bel ^{|p} \oplus Bel ^{\downarrow p})^{\downarrow
X} $$
let us consider the following belief distribution in variables X,Z
\begin{center}
\begin{tabular}{|l|l|r|}
\hline
X & Z & frequencies\\
\hline
\{$x_1$\} & \{$z_1\}$ & 40\\
\{$x_1,x_2$\} & \{$z_2\}$ & 60\\
\hline
\end{tabular}
\end{center}
It is easily to show that marginally consistent $Bel ^{|X}$ does not exist.

What is then the class of belief functions possessing marginally consistent
conditionals~? What is the class of   marginally consistent
conditional belief functions ?

It is easy to show that if there exists a
marginally consistent 
conditional belief function   $Bel
^{|p}$ of the belief function $Bel$ given set of variables $p$ 
then there exists another marginally consistent 
conditional belief function   $Bel_\Xi
^{|p}$ of the belief function $Bel$ given set of variables $p$ such that
$(Bel_\Xi ^{|p})^{\downarrow p}$ has only one focal point 
$(m_\Xi ^{|p})^{\downarrow p}(\Xi_p)=1$ with $\Xi_p$ being the joint domain of
variables from set $p$. In other words, Cano et al. apriori-conditionals 
represent completely the  class of   marginally consistent
conditional belief functions.

Hence it is nearly obvious that in general case-based (separately measured)
belief functions do not possess   marginally consistent
conditional belief functions.

Therefore it is in general of primary interest to find an appropriate
approximation of general belief functions by means of decompositions into
Cano's et al apriori-conditionals. Let us say that an approximation $Bel'$ of
belief function $Bel$ is correct iff for every set A $Bel'(A)<Bel(A)$. An
approximation $Bel'$ of
belief function $Bel$ is marginally correct iff  for every variable X and
every set A ${Bel'} ^{\downarrow X} (A))<Bel ^{\downarrow X}(A)$. 

Let us consider the following algorithm for calculation of a correct 
approximation of the function $Bel$ in variables X,Y:
\begin{itemize}
\item[0.] We initialize the basic probability assignment function $m_{cond}$
defined over variables X,Y with 0 for every subset of $\Xi_X\times \Xi_Y$.
\item[1.] For each set $A_X\subseteq \Xi_X$ with $m ^{\downarrow X}(A_X)>0$
we calculate the quantity $g(A_X,A_Y):=m(A_X\times A_Y)/m ^{\downarrow
X}(A_X)$.
\item[2.] i:=1, q:=0
\item[3.] For each set $A_X\subseteq \Xi_X$ with $m ^{\downarrow X}(A_X)>0$
we select a set $r_i(A_X)\subseteq \Xi_Y$ with  $g(A_X,r_i(A_X))>0$. \\
 $g_{i,r,min}$ be the minimum of $g(A_X,r_i(A_X))$ over all  $A_X\subseteq
\Xi_X$ with $m ^{\downarrow X}(A_X)>0$
\item[4.] For the relation $r_i$ we select a function $a_i:\Xi_X \rightarrow
2^{\Xi_Y}$
such that for each set $A_X\subseteq \Xi_X$ with $m ^{\downarrow X}(A_X)>0$
we have $r_i(A_X)\subseteq \bigcup_{\xi_X\in A_X} a_i(\xi_X)$
\item[5.] We update the function $m_{cond}$ as follows: We calculate the set
$A= \bigcup_{\xi_X\in \Xi_X} \{\xi_X\}\times a_i(\xi_X)$. Then
$m_{cond}(A):= m_{cond}(A)+g_{i,r,min}$.
\item[6.] We update the $g$ function as follows:
For each set $A_X\subseteq \Xi_X$ with $m ^{\downarrow X}(A_X)>0$
$g(A_X,r_i(A_X)):=g(A_X,
r_i(A_X))-g_{i,r,min} $ 
\item[7.] For each set $A_X\subseteq \Xi_X$ with $m ^{\downarrow X}(A_X)>0$
 $$q:=\frac{q+g_{i,r,min}*m ^{\downarrow
X}(A_X)*card(A_X)}{card(\bigcup_{\xi_X\in A_X} a_i(\xi_X))}$$
\item[8.] i:=i+1. If $g$ is equal zero everywhere then terminate, otherwise
continue with step 3.
\end{itemize}
The marginal quality of an approximation constructed by the above
algorithm be the quantity q. The quality q can range from zero to one. If the
quality is equal one then we have constructed a marginally consistent
conditional of Bel given X. If Bel is a probabilistic belief function (with
focal points being singleton sets), then a marginally consistent Cano's
conditional of Bel given X always exists. If Bel is a general belief function
possessing a  marginally consistent Cano's
conditional of Bel given X, then the construction by the above algorithm of
the conditional is characterized by the fact that in step 4 we have
$r_i(A_X)= \bigcup_{\xi_X\in A_X} a_i(\xi_X)$. However, the construction task
as such is hard. In particular, we cannot assume that if there exists a Cano's
conditional of Bel given X, and if and if for i=1,...,k we managed to obtain 
$r_i(A_X)= \bigcup_{\xi_X\in A_X} a_i(\xi_X)$, then we will get a $r_i(A_X)=
\bigcup_{\xi_X\in A_X} a_i(\xi_X)$ for i=k+1. Consider the following
(counter)example: 
Let $Bel$ be a belief function separately measurable in X,Y, marginally
consistent with  $Bel_1 \oplus Bel_2$
where $Bel_1$ being a belief function in X, $Bel_2$ in X,Y with focal 
points:
\begin{center}
\begin{tabular}{|c|r|}
\hline
$A_X$ &  $m_1(A_X)$\\
\hline
\{$x_1$\} & $p_1$\\
\{$x_2$\} & $p_2$\\
\{$x_3$\} & $p_2$\\
\{$x_1,x_2$\} & $p_{12}$\\
\{$x_1,x_3$\} & $p_{13}$\\
\{$x_2,x_3$\} & $p_{23}$\\
\hline
\end{tabular}
\quad \quad %
\begin{tabular}{|c|r|}
\hline
$A_{X,Y}$ &  $m_2(A_{X,Y})$\\
\hline
\{$(x_1,y_1),(x_2,y_1)(x_3,y_2)$\} & $1/3$\\
\{$(x_1,y_1),(x_2,y_2)(x_3,y_2)$\} & $1/3$\\
\{$(x_1,y_2),(x_2,y_1)(x_3,y_2)$\} & $1/3$\\
\hline
\end{tabular}
\end{center}
Obviously  then  $Bel_2$  is  the  marginally   consistent   Cano 
conditional of $Bel$. 
Let us select $r_1$ as follows: 
\begin{center}
\begin{tabular}{|c|r|}
\hline
$A_X$ &  $r_1(A_X)$\\
\hline
\{$x_1$\} & \{$y_1$\}\\
\{$x_2$\} & \{$y_1$\}\\
\{$x_3$\} & \{$y_1$\}\\
\{$x_1,x_2$\} & \{$y_1$\}\\
\{$x_1,x_3$\} & \{$y_1$\}\\
\{$x_2,x_3$\} & \{$y_1$\}\\
\hline
\end{tabular}
\end{center}
We can then easily construct function $a_1$ as $a_1(x_1)=y_1$,  
$a_1(x_2)=y_1$,
 $a_1(x_3)=y_1$ so that  obviously  $r_1(A_X)=
\bigcup_{\xi_X\in A_X} a_1(\xi_X)$. However, if we increase in step 8 i to
i=2 and reenter step 3, then it will not be possible any more to construct
such an $a_2$ that  $r_2(A_X)=
\bigcup_{\xi_X\in A_X} a_2(\xi_X)$ because necessarily a fragment of $r_2$
will be:
\begin{center}
\begin{tabular}{|c|r|}
\hline
$A_X$ &  $r_2(A_X)$\\
\hline
\dots     & \dots\\
\{$x_1,x_2$\} & \{$y_1,y_2$\}\\
\{$x_1,x_3$\} & \{$y_1,y_2$\}\\
\{$x_2,x_3$\} & \{$y_1,y_2$\}\\
\hline
\end{tabular}
\end{center}
Hence in general finding a marginally consistent Cano's conditional, even if
it exists, is hard and requires backtracking. So probably one will be
satisfied already if one finds a  high quality marginally correct approximate
conditional belief distribution (Application of genetic algorithms is
advised here.).

\section{IMPACT ON REASONING}

Let $Bel$ be a belief function in X,Y and $Bel ^{|X}$ its marginally correct
approximate conditional belief distribution. 
Then $Bel ^{\downarrow X}\oplus Bel ^{|X}$ is a marginally correct
approximation of $Bel$. Then Shafer's conditioning $Bel(|| A_X \times \Xi_Y)$
with $A_X$ being any non-empty subset of the domain $\Xi_X$ of $X$ 
is marginally correctly approximated by 
$(Bel ^{\downarrow X}\oplus Bel ^{|X})(||  A_X \times \Xi_Y)$, that is given
the "decomposition" of $Bel$ into its marginal on X and the approximately
correct conditional on X we can reason approximately correctly about
the a posteriori distribution of
Y given any event of  observation of the intrinsic value of X. However,
we cannot do it in the reverse direction: we  cannot  derive  the 
aposterioric
distribution of X by observation of Y because in general  $Bel(|| \Xi_X
\times
A_Y)$ with $A_Y$ being any non-empty subset of the domain $\Xi_Y$ of $Y$ 
is NOT marginally correctly approximated by 
$(Bel ^{\downarrow X}\oplus Bel ^{|X})(||  \Xi_X \times A_Y)$. What is worse,
even if the conditionals are marginally consistent, we will not achieve 
marginal correctnesses, not to say marginal consistency.

Thus,
if we have a marginally consistent factorization
 of a belief
function then we can in general use neither Shafer \& Shenoy nor Cano's at al.
framework for propagation of uncertainty because the results will be
inconsistent with the data. 
A way out of this problem, for the framework of Cano, is to consider separate
marginally correct approximations in each direction of reasoning. 
Given the polytree representing correctly dependencies among
variables, we need to transform this polytree, for every variable the value
of which we want to infer, into a such a (belief) network graph that every
arrow points towards that variable. In case of plain directed trees the result
of this transformation is a polytree (with undirected backbone identical
with that of the original tree) such that edges connecting the target
variable with its neighbous are reoriented to point at the target variable and
every other variable is connected via a directed path with the target
variable. In case of general polytree the result of the transformation is a
network with more edges than the original polytree. Edges are added when an
edge is reversed which originally pointed at a node with many ingoing edges.
E.g. if we had the situation that $X->Z<-Y$ and we want to invert the edge
$Z<-Y$, then we need to add the edge $X->Y$. In case of general polytree
we obtain in this way networks with considerably different undirected
backbones\footnote{In case of a general belief network transformations aiming
at creation of directed pathes from every node towards the target variable
would result in still more complicated networks so that merits of reasoning in
such networks would be questionable}. 

If we assume that we adopt a structure     (factorization) of Cano type, that
is in form of a generalized polytree (singly connected "bayesian" network),
then 
products of above transformations
are               directly convertible into a Shenoy's and Shafer's
Markov tree.            Let us now shift to Shenoy's and Shafer's belief
propagation in Markov tree: The general principle there is "message passing" -
if a node of Markov-tree gets information from its all but one  neighbors ,
then it sends, to the remaining node, a "message", that is $\oplus$
combination
of those messages plus its own factor of the belief function factorization. In
the original Shenoy/Shafer algorithm, this node's own factor of the belief
function factorization is exactly the same independly to which neighbor the
message is sent. We propose to have separate hypertrees for each target
variable and to reason within each of them in one direction only (resp.
modifications of propagation algorithm are known). 
Then it is guaranteed that the results of reasoning (a posteriori marginal
distributions) will be  marginally  correct  approximations  with 
respect to the
intrinsic distribution.

Cano et al. conditionals are  in fact
sets of mappings between sets of variables selected with some probability.
This gives a new meaning to the belief function. Instead of thinking in the
way probability functions do that is that given some value of one 
variable, the
conditional probability distribution assigns a value to another variable,
we can think of objects that are assigned with some probability a belief.

 {
\section{DISCUSSION}

Smets \cite{Smets:92} stated that domains of MTE
applications
are those where "we are ignorant of the existence of probabilities", 
and warns that MTE is 
"not a 
model for poorly known probabilities" (\cite{Smets:92}, p.324). Smets states
further "Far too often, authors concentrate on the static component (how
beliefs are 
 allocated?) and discover many relations between TBM (transferable belief 
model of Smets) 
 and ULP (upper lower probability) models, inner and outer measures 
(Fagin and Halpern \cite{Fagin:91}), random sets (Nguyen \cite{Nguyen:78}), 
probabilities of provability 
 (Pearl \cite{Pearl:88}), probabilities of necessity (Ruspini 
\cite{Ruspini:86}) etc. But these authors 
usually do not explain or justify the dynamic component (how are beliefs 
updated?), that  is, how updating (conditioning) is to be handled (except in 
some cases by defining conditioning as a special case of combination). So I 
 (that is Smets) feel that these partial comparisons are incomplete, 
especially 
as all these interpretations lead to different updating rules." 
(\cite{Smets:92}, pp. 324-325).  Ironically, Smets gives later
in the same paper
 an example of
belief function ("hostile-Mother-Nature-Example") which may be clearly
considered as lower probability interpretation of belief function, just, at
further consideration, leading to very same pitfalls as approaches criticized
himself. 

In order to explicate the reasons of failures of various attempts 
to establish a case-based interpretation of MTE, 
in this paper we drew our attention to the questions related to  updating
(conditioning) in this theory.
We have investigated some fundamental problems of case-based
interpretation of the Dempster-Shafer Mathematical Theory of Evidence. 
\begin{itemize}
\item The problem of the meaning of a case for MTE. It has been argued that if
some observed attributes are set-valued then the meaning of the case should be
the cross-product of all these sets (and not any subset, e.g. a marginally
consistent subset of this cross product).
\item The problem of meaning of aposteriori distribution  given observation of
some variables (Smets; "updating" problem). It has been argued that, for
calculation of empirical
(case-based) MTE aposteriori belief distributions, the interesting cases need
not only to be selected but also updated
\item The problem of apriori-conditioning: it has been argued that in general
apriori conditionals of (case-based) belief functions are not belief 
functions,
 but rather pseudo-belief functions (with negative basic "probability"
assigments), hence lacking a case-baaed interpretation. 
What is more, even every belief function marginally consistent
with a given case-based belief distribution may lack  apriori conditional
belief function. In the rare cases when a marginally consistent belief
function has an apriori conditional  belief function, there exits 
always Cano's
conditional belief function for that belief function. 
\item The problem of marginally correct approximation of a belief function. It
has been argued that even if a marginally consistent  Cano's
conditional belief function for a belief function exists, it is hard to find.
Therefore one needs to look for algorithms constructing  marginally correct
approximated apriori conditional  belief functions. A general frame algorithm
has been given together with a proposed function of quality evaluation of the
approximation. 
\item The problem of reasoning with marginally correct approximations via
conditional mar\-gin\-al\-ly correct approximated apriori belief functions. It
has been argued that, if we want to achieve results corresponding properly to
empirical (case-based) belief functions, it is necessary to abandon the
traditional form of uncertainty propagation proposed by Cano et al and by
Shenoy and Shafer. Instead of Markov fields, which are insensitive to the
orientation of reasoning, one shall use unidirectional  mode of reasoning.
\end{itemize}
}
\section{CONCLUSION}

\begin{enumerate}
\item The case-based derivation of aposteriori MTE belief function requires
combination of operations of case selection with operation of case
updating.
\item Therefore,
marginally correct inference rules for MTE derived from data can be
applied only unidirectionally - reversal of direction of reasoning leads to
contradictions with the data.  For this reason, original Cano et al. and
Shenoy/Shafer
uncertainty propagation methods are not suitable for case-based MTE belief
functions.
\item Hence in case of direct dependence of a set of n variables, n inferences
networks - one for each variable as dependent on the remaining ones -
have to be established and used depending on target variable.
\end{enumerate}

\end{document}